\documentclass{article}
\usepackage{amsmath,graphicx,mlspconf}
\usepackage{amsfonts}
\usepackage{xcolor}
\usepackage{amssymb}
\usepackage{subcaption}
\usepackage{float}

\copyrightnotice{979-8-3503-2411-2/25/\$31.00 {\copyright}2025 IEEE}

\toappear{2025 IEEE International Workshop on Machine Learning for Signal Processing, Aug.\ 31-- Sep.\ 3, 2025, Istanbul, Turkey}

\title{Benefits of Online Tilted Empirical Risk Minimization:\\ A Case Study of Outlier Detection and Robust Regression}

\name{Yiğit E. Yıldırım,\; Samet Demir,\; Zafer Doğan\thanks{We acknowledge that this work is supported partially by TÜBİTAK under project 124E063 in ARDEB 1001 program. Y.E.Y. is supported by the TÜBİTAK project. S.D. is supported by an AI Fellowship provided by KUIS AI Research Center and a PhD Scholarship (BİDEB 2211) from TÜBİTAK. The corresponding author is Zafer Doğan (zdogan@ku.edu.tr).}}
\address{
    MLIP Research Group, KUIS AI Center \& Department of EEE, Koç University\\
    İstanbul, Turkey
}

\begin{document}

\maketitle

\begin{abstract}
    Empirical Risk Minimization (ERM) is a foundational framework for supervised learning but primarily optimizes average-case performance, often neglecting fairness and robustness considerations. Tilted Empirical Risk Minimization (TERM) extends ERM by introducing an exponential tilt hyperparameter $t$ to balance average-case accuracy with worst-case fairness and robustness. However, in online or streaming settings where data arrive one sample at a time, the classical TERM objective degenerates to standard ERM, losing tilt sensitivity. We address this limitation by proposing an online TERM formulation that removes the logarithm from the classical objective, preserving tilt effects without additional computational or memory overhead. This formulation enables a continuous trade-off controlled by $t$, smoothly interpolating between ERM ($t \to 0$), fairness emphasis ($t > 0$), and robustness to outliers ($t < 0$). We empirically validate online TERM on two representative streaming tasks: robust linear regression with adversarial outliers and minority-class detection in binary classification. Our results demonstrate that negative tilting effectively suppresses outlier influence, while positive tilting improves recall with minimal impact on precision, all at per-sample computational cost equivalent to ERM. Online TERM thus recovers the full robustness-fairness spectrum of classical TERM in an efficient single-sample learning regime.
\end{abstract}


\begin{figure*}[htp]
    \begin{subfigure}[t]{0.28\textwidth}
        \centering
        \includegraphics[width=\linewidth]{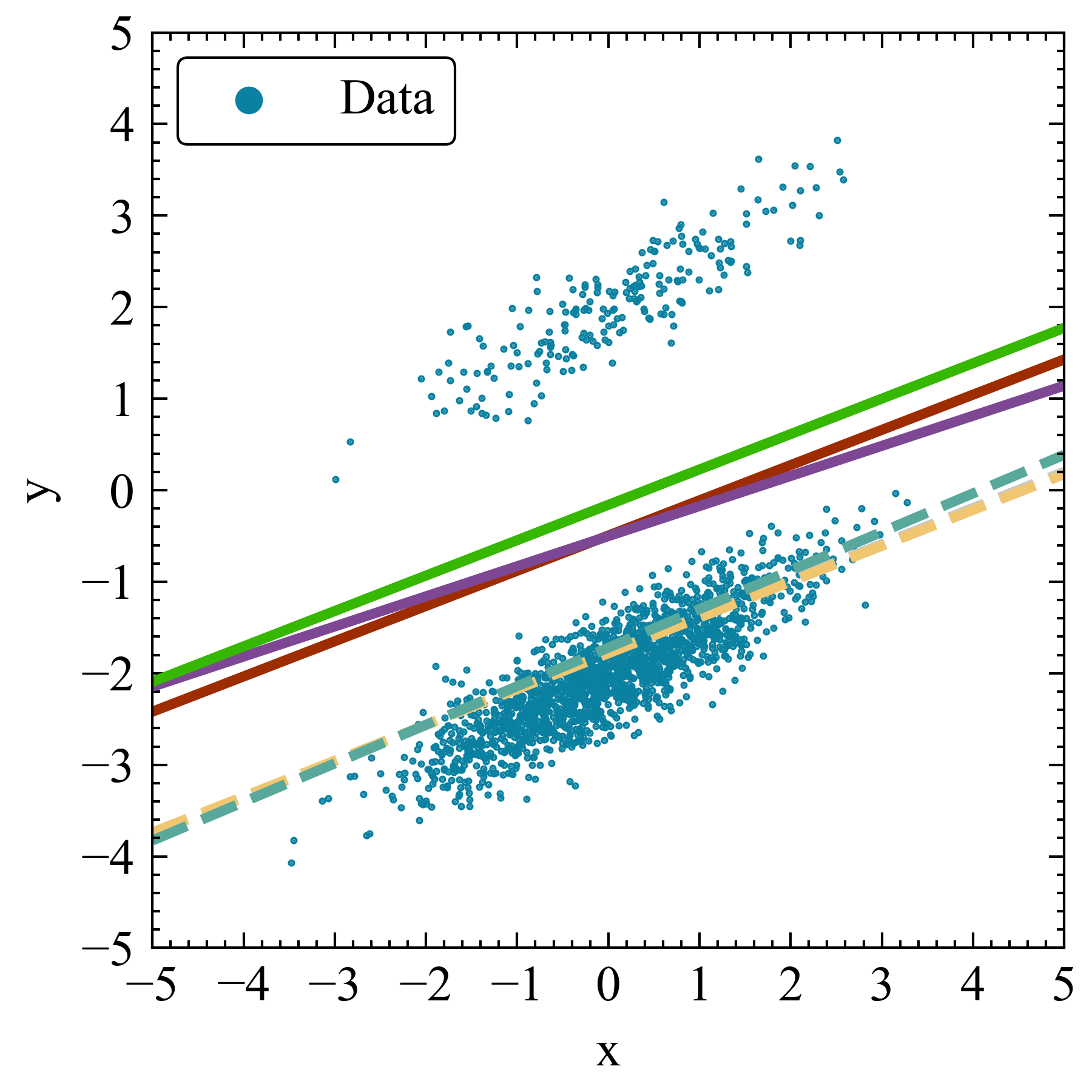}
        \caption{Online TERM with $t = +0.5$: fitted lines emphasize large-loss samples.}
        \label{fig:1}
    \end{subfigure}
    \hfill
    \begin{subfigure}[t]{0.28\textwidth}
        \centering
        \includegraphics[width=\linewidth]{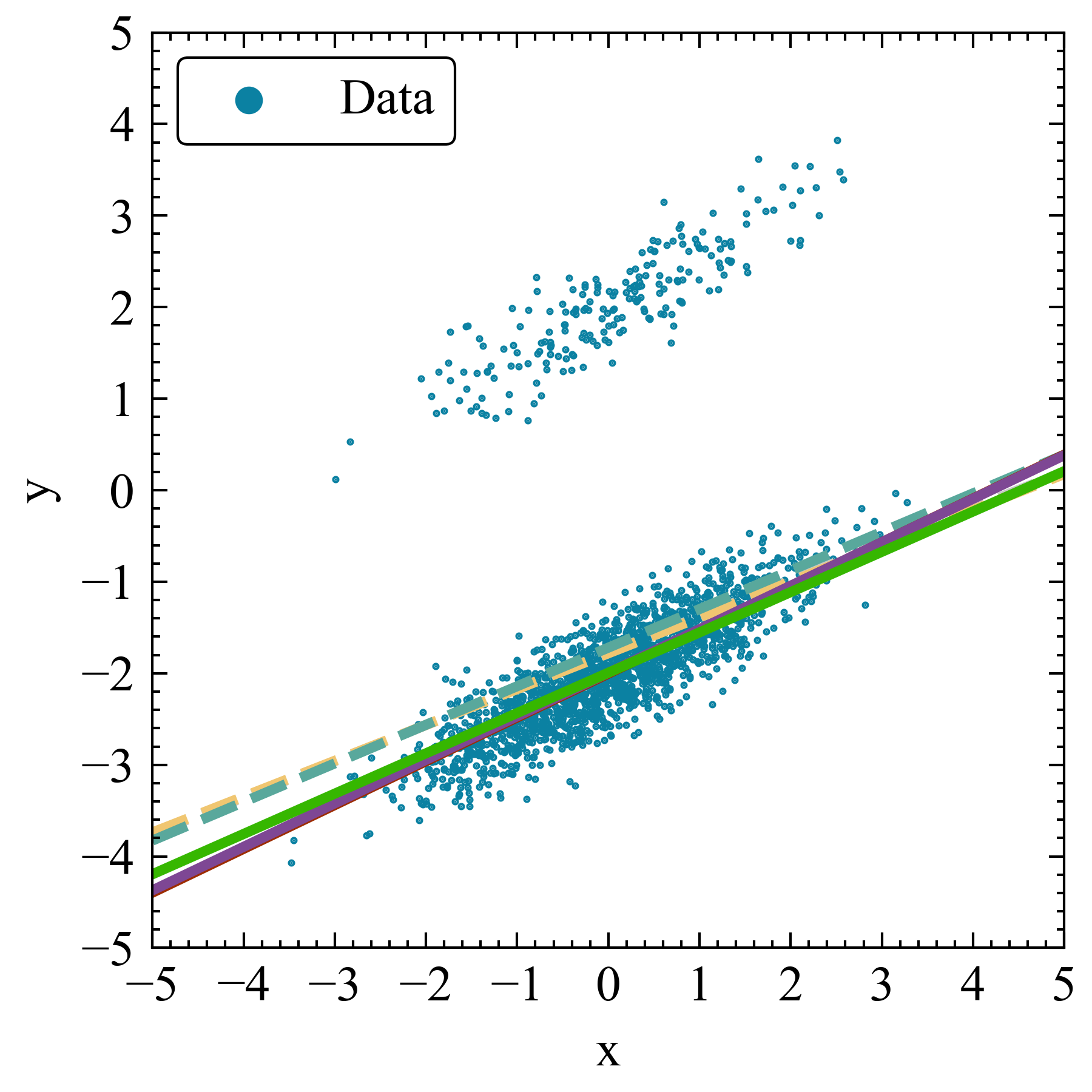}
        \caption{Online TERM with $t = -0.5$: fitted lines downweight outlier influence.}
        \label{fig:2}
    \end{subfigure}
    \hfill
    \begin{subfigure}[t]{0.41\textwidth}
        \centering
        \includegraphics[width=\linewidth]{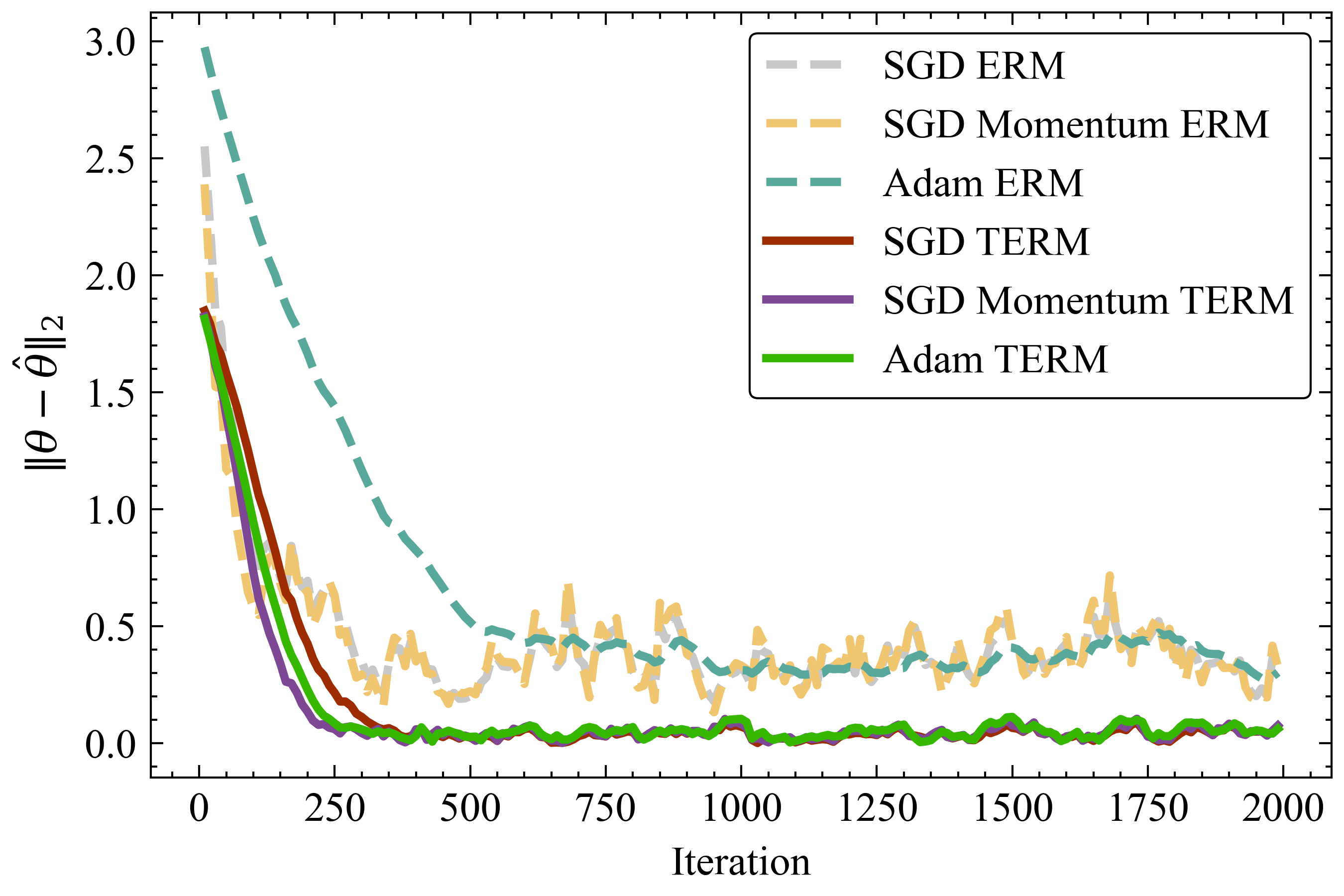}
        \caption{Prediction error with $t < 0$: Euclidean distance to true inlier weights, recorded every 10 iterations.}
        \label{fig:3}
    \end{subfigure}
    \caption{We illustrate a toy example of noisy linear regression with outliers to highlight the differences between ordinary ERM and the online $t$‑tilted objective. (a) and (b) compare both approaches on a synthetic stream of samples, where $90\%$ of the points follow the inlier line $y = 0.52x - 2$ and the remaining $10\%$ are parallel outliers shifted upward by four units, $y = 0.52x + 2$. Each observation is perturbed by i.i.d. Gaussian noise with standard deviation $0.3$. The squared loss is used while models are updated one sample at a time—without batching—using various optimizers. Learning rates are set to $10^{-2}$ for ERM, $10^{-2}$ for TERM with $t < 0$, and $10^{-4}$ for TERM with $t > 0$. SGD Momentum ERM/TERM uses the momentum parameter of 0.3, and hyperparameters of Adam ERM/TERM are set to $\beta_1=0.9,\beta_2=0.999,\epsilon=10^{-8},\overline{\epsilon}=0$. These rates are chosen to balance stability and to reveal long-term behavior. (c) tracks the Euclidean distance $\|\boldsymbol{\theta} - \boldsymbol{\theta}^*\|_2$—computed every ten updates—between the current parameter vector and the ground truth inlier parameters $\boldsymbol{\theta}^* = (0.52, -2)$. }
    \label{fig:robust_regression}
\end{figure*}

\section{Introduction}
\label{sec:intro}
Empirical Risk Minimization (ERM) is a widely used principle underlying most modern supervised learning algorithms \cite{goodfellow2016book}. The ERM objective selects model parameters $\boldsymbol{\theta}$ by minimizing the average loss over $N$ samples:
\begin{equation}
R(\boldsymbol{\theta}) := \frac{1}{N} \sum_{i=1}^N \ell_i(\boldsymbol{\theta}),
\end{equation}
where $\ell_i$ denotes the loss for the $i$-th sample.

ERM enjoys elegant statistical guarantees and provides strong predictive performance when the training data faithfully represent the test distribution. However, by design, ERM focuses on mean performance. Specifically, every example contributes equally to the objective, regardless of whether it is representative, atypical, or even incorrectly labeled. This uniform weighting makes ERM fragile in the presence of outliers, noisy data \cite{pmlr-v80-jiang18c, Learning-From-Noisy-Singly-labeled-Data}, or class imbalance \cite{Lin_2017_ICCV, 6126229}, all of which are common in practical learning scenarios. A substantial body of work has addressed these shortcomings of standard ERM \cite{pmlr-v80-hashimoto18a, NEURIPS2018_cc4af25f, JMLR:v20:17-750}. Researchers have proposed fairness-aware objectives that ensure balanced training across subgroups \cite{pmlr-v80-hashimoto18a, NEURIPS2018_cc4af25f} and robust formulations that improve test performance \cite{JMLR:v20:17-750}. In the search for alternatives to ERM, various tilting techniques have been explored across different settings, including importance sampling \cite{1459045}, sequential decision making \cite{Howard1972, 8967699}, and large deviation theory \cite{8522043}. Tilted Empirical Risk Minimization (TERM) extends the ERM framework with a tilt hyperparameter \cite{DBLP:journals/corr/abs-2007-01162, JMLR:v24:21-1095, aminian2024generalization}. Instead of minimizing the average loss, TERM minimizes a tilted loss:
\begin{equation}
\bar{R}(t; \boldsymbol{\theta}) := \frac{1}{t} \log\left( \frac{1}{N} \sum_{i=1}^N e^{t \ell_i(\boldsymbol{\theta})} \right),
\label{eq:batch_term}
\end{equation}
where $t \in \mathbb{R}$ is the tilting hyperparameter. The objective reduces to ordinary ERM as $t \to 0$. For $t > 0$, the exponential weighting magnifies large individual losses, smoothly interpolating toward the max-loss objective \cite{Pee2010, 4044894}, thus promoting fairness. For $t < 0$, the weighting suppresses large losses, making the objective robust to corrupt or adversarial samples \cite{Wang01062013}. Therefore, the tilting hyperparameter $t$ provides a continuous spectrum of trade-offs between fairness and robustness.

TERM is typically formulated for batch learning, where losses are aggregated over a set of $N > 1$ samples before a parameter update. However, in online settings—where the learner processes one sample per iteration (i.e., $N = 1$)—the TERM objective \eqref{eq:batch_term} collapses to ordinary ERM, with the tilt hyperparameter $t$ effectively disappearing. This leads to the loss of TERM's fairness and robustness benefits.

To address this limitation, we propose an online TERM formulation by introducing a simple modification that preserves the tilt hyperparameter in the online regime. We then systematically demonstrate the practical benefits of this new formulation. Specifically, we conduct a study on two toy problem settings: robust linear regression and outlier detection in binary classification. In both cases, data generation includes controlled outlier samples. By varying the tilt hyperparameter $t$ across positive and negative ranges, we quantify how online TERM trades off robustness and fairness. These controlled experiments demonstrate that a single hyperparameter $t$ can smoothly interpolate between average loss, fairness, and robustness. Hence, we empirically show that online TERM is a practical tool for problems involving fairness and robustness concerns.

\begin{figure*}[!htp]
    \begin{subfigure}[t]{0.28\textwidth}
        \centering
        \includegraphics[width=\linewidth]{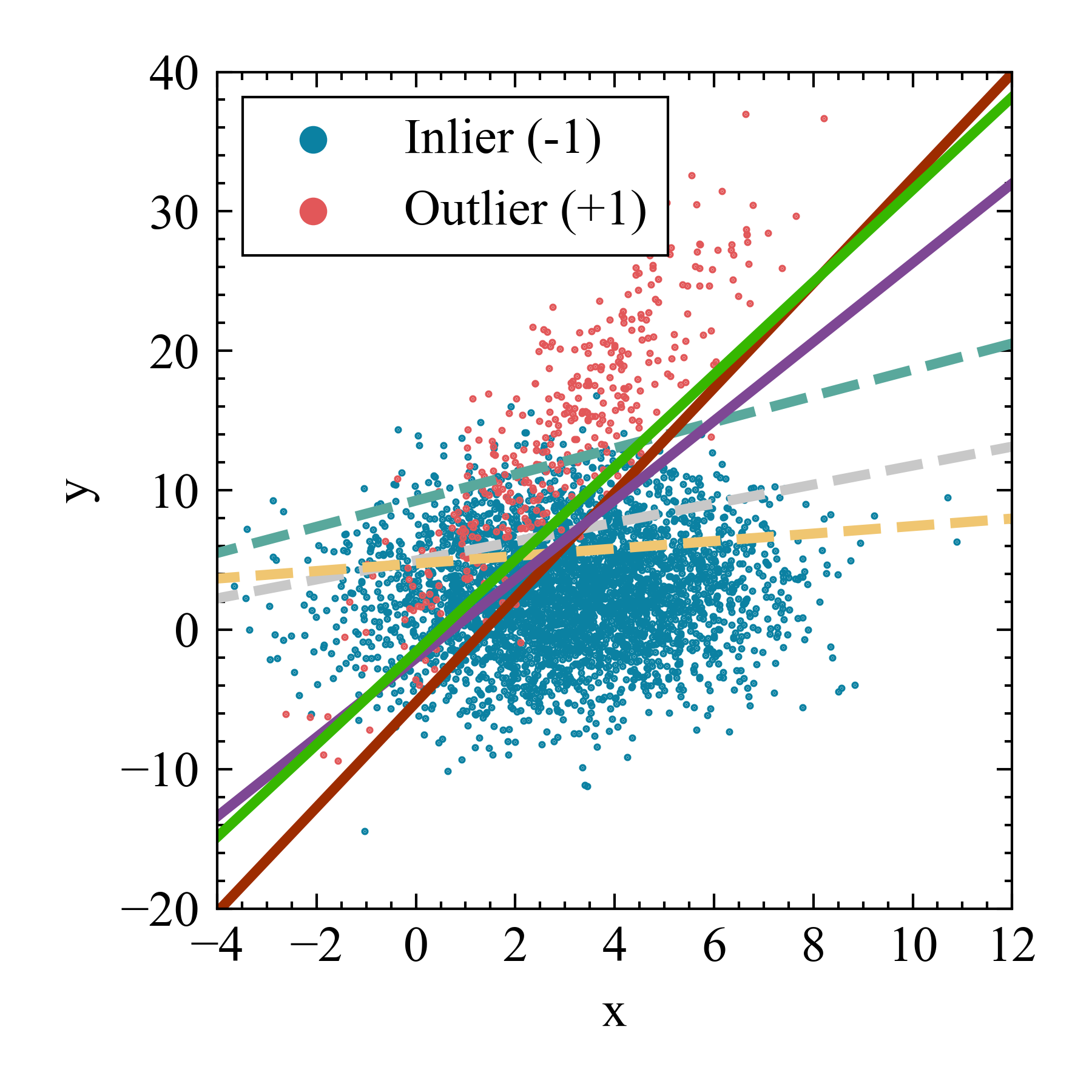}
        \caption{Decision boundaries obtained with online TERM using tilt $t = 0.2$.}
        \label{fig:4}
    \end{subfigure}
    \begin{subfigure}[t]{0.28\textwidth}
        \centering
        \includegraphics[width=\linewidth]{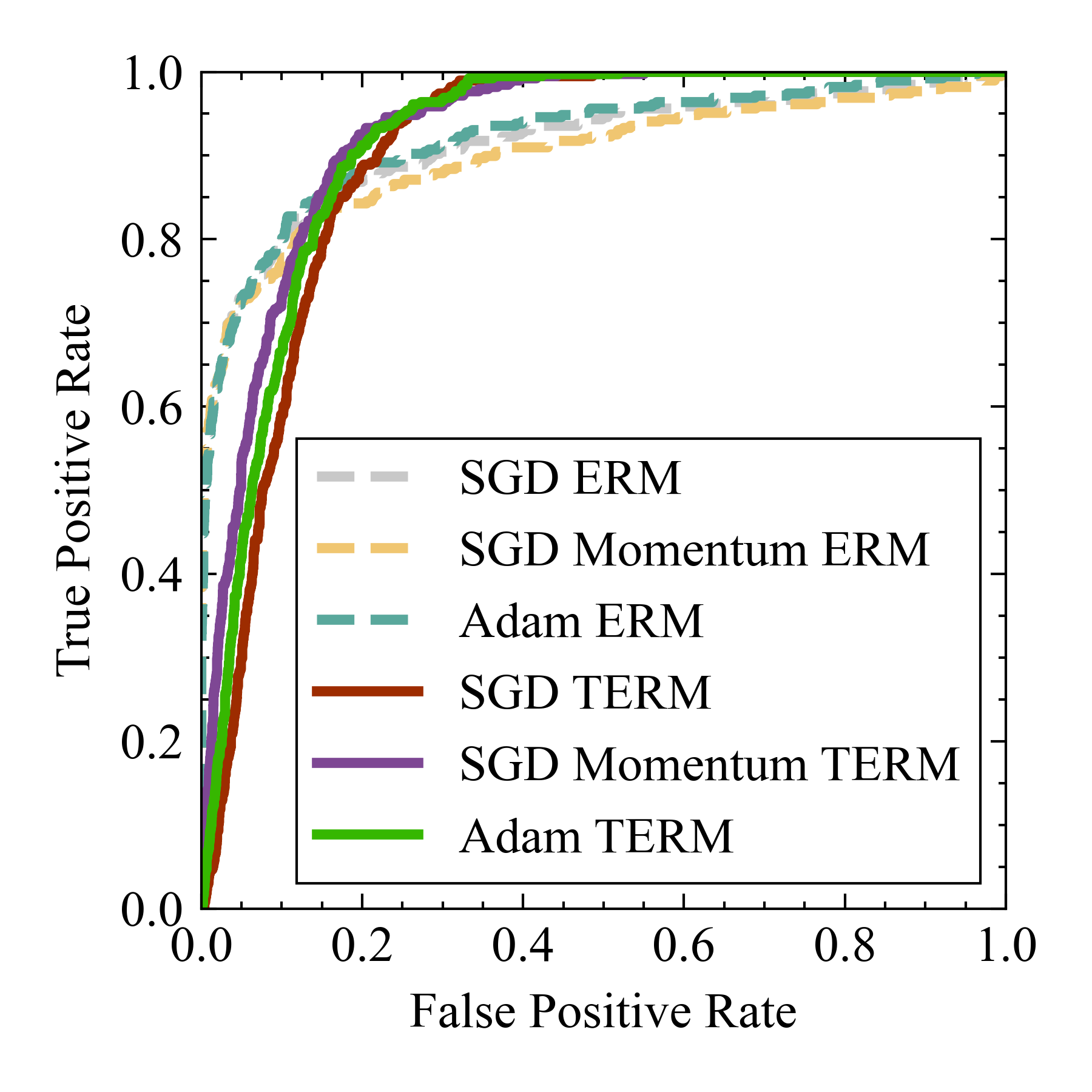}
        \caption{ROC curves obtained with online TERM using tilt $t = 0.2$.}
        \label{fig:5}
    \end{subfigure}
    \begin{subfigure}[t]{0.40\textwidth}
        \centering
        \includegraphics[width=\linewidth]{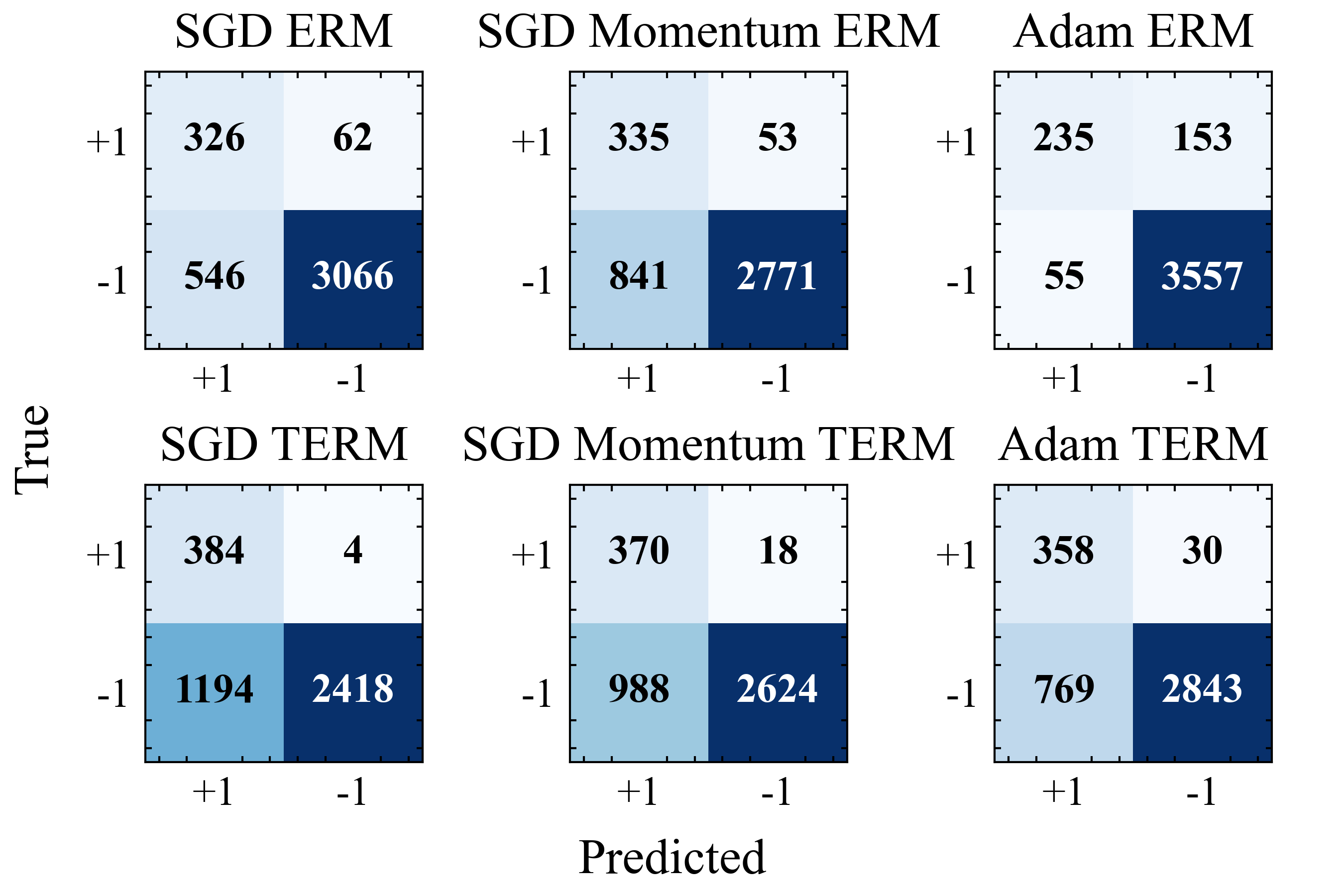}
        \caption{Confusion matrices obtained with online TERM using tilt $t = 0.2$.}
        \label{fig:6}
    \end{subfigure}
    \caption{We illustrate a toy binary classification task with inliers and outliers to study the behavior of online TERM under positive tilting. At each iteration, the learner observes a single sample $(x, y)$ drawn independently from a mixture of two linear models with additive Gaussian noise. With probability $0.9$, the inlier samples are drawn according to $y = 0.2x + 2 + \varepsilon, \; \varepsilon \sim \mathcal{N}(0, \sigma^2),$ and are assigned the label $-1$. With probability $0.1$, the outlier samples are drawn according to $y = 4x + 2 + \varepsilon, \; \varepsilon \sim \mathcal{N}(0, \sigma^2),$ and are assigned the label $+1$. The noise standard deviation is set to $\sigma = 4.0$ to simulate high-variance perturbations. The learner maintains a linear score function of the form $s_{\boldsymbol{\theta}}(x, y) = \boldsymbol{\theta}^\top (x, y, 1)$, where $\boldsymbol{\theta} \in \mathbb{R}^3$, and updates the parameters over 4000 iterations using plain SGD, SGD with momentum, or Adam optimizers together with squared loss. The SGD ERM, SGD Momentum ERM, and Adam ERM variants use learning rates of $10^{-3}$, $10^{-3}$, and $2 \times 10^{-3}$, respectively. The corresponding TERM variants are trained with reduced learning rates—$2 \times 10^{-4}$ for SGD TERM and SGD Momentum TERM, and $8 \times 10^{-4}$ for Adam TERM—to counteract the risk of exploding gradients induced by the positive tilt. SGD Momentum ERM/TERM uses the momentum value of 0.3, and hyperparameters of Adam ERM/TERM are set to $\beta_1=0.9,\beta_2=0.999,\epsilon=10^{-8},\overline{\epsilon}=0$. }
    \label{fig:classification_positive_tild}
\end{figure*}

\section{Online Tilted Empirical Risk Minimization}

In this section, we derive an online version of TERM that preserves the tilting hyperparameter. To this end, we begin by examining the gradient of the tilted objective \eqref{eq:batch_term}:
\begin{equation}
\nabla_{\boldsymbol{\theta}}\bar{R}(t; \boldsymbol{\theta}) = \sum_i \frac{e^{t \ell_i(\boldsymbol{\theta})}}{\sum_j e^{t \ell_j(\boldsymbol{\theta})}} \nabla_{\boldsymbol{\theta}} \ell_i(\boldsymbol{\theta}),
\end{equation}
where each example influences the gradient proportionally to the normalized weight $e^{t \ell_i(\boldsymbol{\theta})} / \sum_j e^{t \ell_j(\boldsymbol{\theta})}$. 

To retain this weighting mechanism in the online setting where $N=1$, we drop the logarithm from \eqref{eq:batch_term} and propose the following alternative tilted objective:
\begin{equation}
\tilde{R}(t; \boldsymbol{\theta}) := \frac{1}{t} e^{t \ell_i(\boldsymbol{\theta})}.
\label{eq:online_term}
\end{equation}
This objective yields the gradient
\begin{equation}
\nabla_{\boldsymbol{\theta}} \tilde{R}(t; \boldsymbol{\theta}) = e^{t \ell_i(\boldsymbol{\theta})} \nabla_{\boldsymbol{\theta}} \ell_i(\boldsymbol{\theta}),
\label{eq:online_term_grad}
\end{equation}
where the exponential factor $e^{t \ell_i(\boldsymbol{\theta})}$ amplifies large loss samples when $t > 0$ and attenuates them when $t < 0$. This exponential weighting reinstates the ability to control the fairness and robustness trade-offs inherent to TERM. 

However, unlike the batch TERM, the exponential factor in \eqref{eq:online_term_grad} may explode for positive $t$, necessitating the careful selection of learning rates. In the remainder of this work, we demonstrate the advantages of employing the online TERM formulation \eqref{eq:online_term} in robust regression and outlier detection tasks.

\subsection{Robust linear regression with online TERM} \label{sec:2}

In this subsection, we present simulation results for robust linear regression using online TERM. Specifically, we consider a linear regression problem in which 90\% of the data are inliers and the remaining 10\% are outliers. In this setting, ERM fails to maintain robustness against the outliers, whereas TERM can effectively handle them when the tilting hyperparameter $t$ is chosen appropriately. 

Figure \ref{fig:robust_regression} illustrates our results on this robust regression problem. In Figure \ref{fig:1}, the tilt hyperparameter is set to $t = 0.5$. Despite differences in their update rules, various optimizers—SGD, SGD with momentum, and Adam—exhibit qualitatively similar behavior: they produce regression lines that are pulled between the inlier and outlier samples. This occurs because the exponential weight $e^{t \ell_i(\boldsymbol{\theta})}$ grows rapidly for large losses, causing the regression line to be influenced heavily by outliers.

Conversely, Figure \ref{fig:2} shows results for $t = -0.5$. Here, the negative tilt suppresses large losses, causing TERM optimizers to assign negligible weight to outliers and thus maintain robustness. The similarity in fitted lines across different optimizers highlights two key observations: (i) negative tilting restores the robustness that the original batch TERM objective typically provides, and (ii) this robustness is consistent across optimizer choices. These findings demonstrate that online TERM with $t < 0$ is beneficial for robust linear regression in the presence of outliers.

Compared to ERM-based optimizers, TERM optimizers with negative tilt ($t < 0$) achieve a better fit to the ground truth, as evident in Figures \ref{fig:2} and \ref{fig:3}. In particular, Figure \ref{fig:3} shows that although all optimizers start from the same random initialization, their parameter trajectories diverge over time when processing the same data stream. The weight errors of TERM optimizers with negative tilt converge more steadily toward zero within the first few hundred iterations and then stabilize. This improved consistency arises from the factor $e^{t \ell_i(\boldsymbol{\theta})}$, which significantly reduces gradient magnitude, thereby limiting oscillations and abrupt updates. In contrast, ERM optimizers ($t=0$) exhibit more oscillatory behavior and poorer convergence.

Importantly, once TERM optimizers with negative tilt approach the true parameters, they remain stable for the duration of training without signs of oscillation. This quantitative evidence complements the visual interpretation from the fitted regression lines and supports the conclusion that modestly negative tilting reliably downweights outliers, accelerates convergence, and maintains parameter stability regardless of the optimizer used.

\begin{figure*}[!htp]
    \begin{subfigure}[t]{0.28\textwidth}
        \centering
        \includegraphics[width=\linewidth]{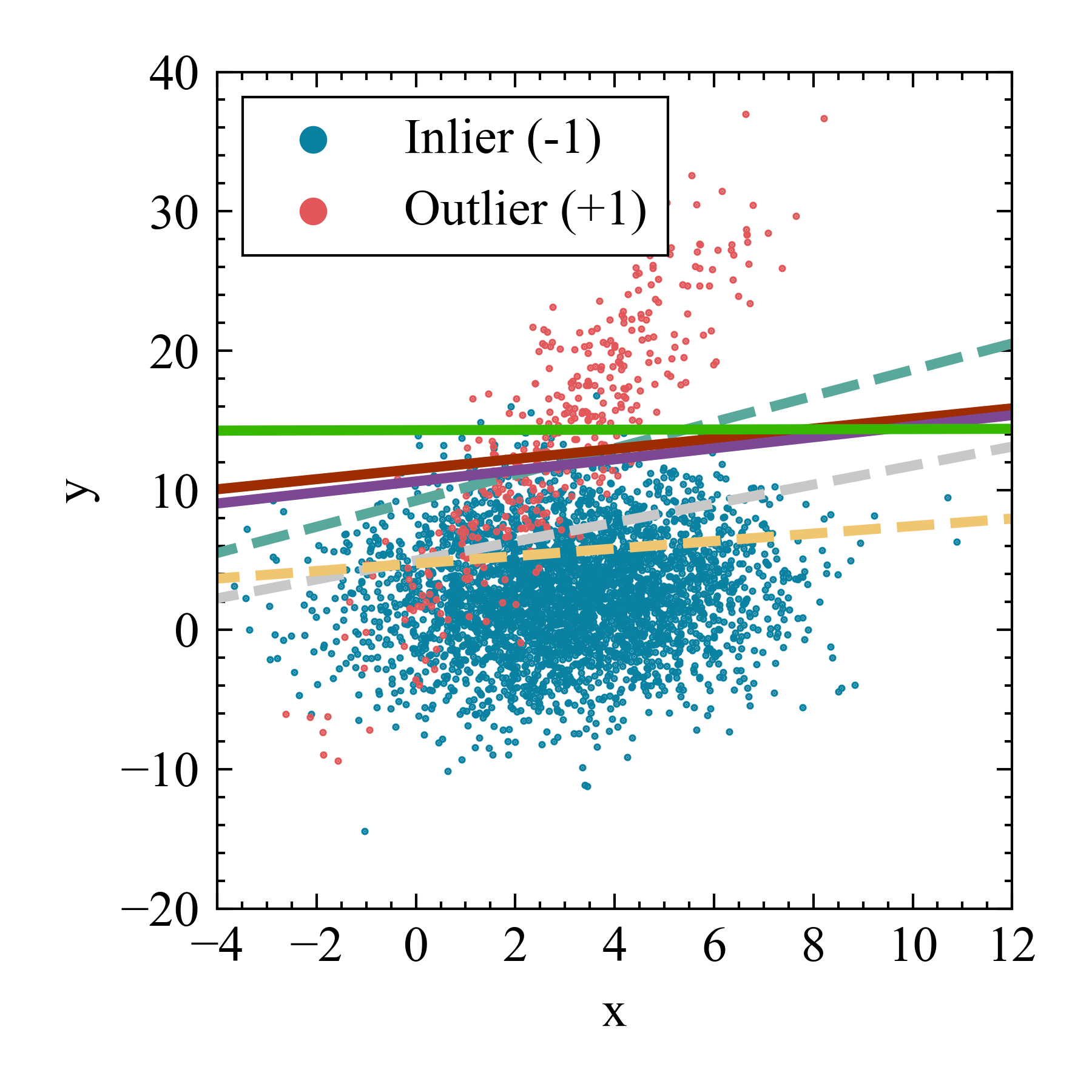}
        \caption{Decision boundaries obtained with online TERM using tilt $t = -0.2$.}
        \label{fig:7}
    \end{subfigure}
    \begin{subfigure}[t]{0.28\textwidth}
        \centering
        \includegraphics[width=\linewidth]{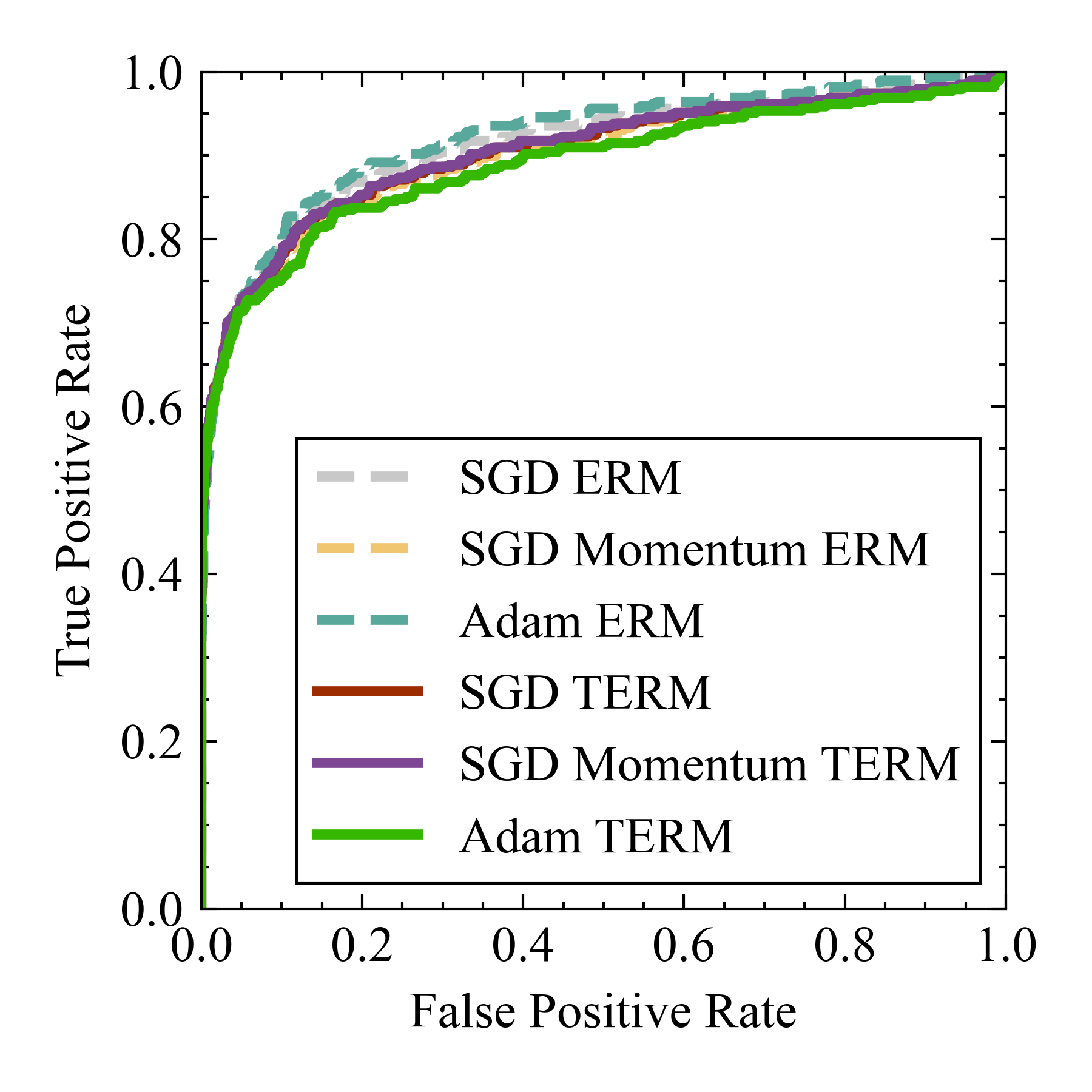}
        \caption{ROC curves obtained with online TERM using tilt $t = -0.2$.}
        \label{fig:8}
    \end{subfigure}
    \begin{subfigure}[t]{0.40\textwidth}
        \centering
        \includegraphics[width=\linewidth]{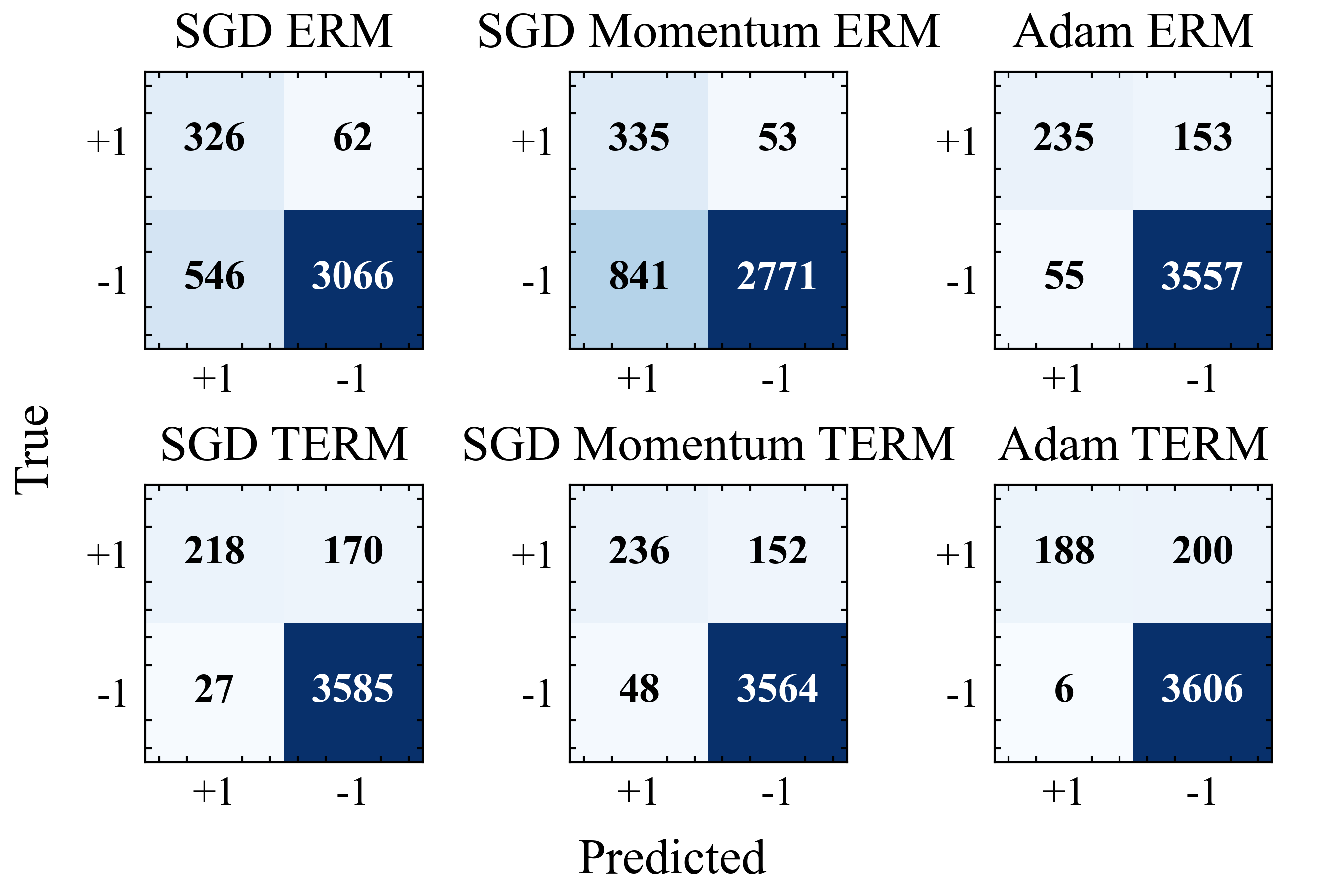}
        \caption{Confusion matrices obtained with online TERM using tilt $t = -0.2$.}
        \label{fig:9}
    \end{subfigure}
    \caption{We illustrate a toy binary classification example with outliers using online TERM under negative tilting ($t < 0$). The setting is the same as Figure \ref{fig:classification_positive_tild} except for the details explained here. At each iteration, the learner receives a single $(x, y)$ pair drawn from a corrupted linear model as previously described. The SGD TERM, SGD Momentum TERM, and Adam TERM optimizers use learning rates of $2 \times 10^{-3}$, $2 \times 10^{-3}$, and $8 \times 10^{-3}$, respectively. These increased learning rates compensate for the diminished gradient magnitudes caused by the negative tilt hyperparameter, which strongly downweights large residuals. This adjustment ensures effective parameter updates despite the suppressed influence of outlier samples.}
    \label{fig:classification_negative_tild}
\end{figure*}

In summary, the online $t$-tilted update with negative tilt offers a lightweight, tuning-free alternative to robust estimators. It preserves the simplicity of the quadratic loss and processes data points sequentially. As demonstrated in Figure \ref{fig:3}, where the weight error curves diminish, it converges rapidly to the true inlier line, while ERM variants are adversely affected by outliers.

\subsection{Outlier detection in binary classification with online TERM} \label{sec:3}

We next apply the online $t$-tilted update to a streaming binary classification task aimed at detecting outliers while minimizing misclassification of these rare points.

In this setting, large loss examples correspond precisely to the outliers we intend to identify. Setting $t = 0.2$ amplifies their influence (see Figure \ref{fig:classification_positive_tild}), as the exponential weight can be orders of magnitude larger for an outlier compared to a typical inlier. Consequently, the decision boundaries in Figure \ref{fig:4} adapt to correctly classify outliers. This effect is evident in the confusion matrix shown in Figure \ref{fig:6}, which reveals an increase in true positives (correctly identified outliers) and a decrease in false negatives (missed outliers), albeit at the expense of more false positives (misclassified inliers). Figure \ref{fig:5} further supports this, with the ROC curves for TERM optimizers rising above those of the ERM baseline, indicating improved recall beyond certain thresholds. When false positives are tolerable, TERM optimizers thus effectively identify nearly all outliers, minimizing false negatives.

Conversely, setting $t = -0.2$ (see Figure \ref{fig:classification_negative_tild}) reduces the gradient magnitude whenever the squared error is large, effectively instructing the optimizer to downweight outliers. As depicted in Figure \ref{fig:7}, this produces a near-horizontal decision boundary that aligns closely with the inlier distribution. The confusion matrix in Figure \ref{fig:9} confirms a significant reduction in false positives and an increase in true negatives, though at the cost of missing approximately half of the outliers. Correspondingly, the ROC curves in Figure \ref{fig:8} demonstrate that TERM optimizers with negative tilt outperform their ERM counterparts in the low false positive rate region, achieving higher true positive rates while maintaining a low false positive count. This makes them preferable when false positives carry a high cost, emphasizing their bias toward robustness over sensitivity.

These findings parallel the observations from the regression experiments in Section \ref{sec:2}. The sign of the tilt parameter $t$ acts as a tunable trade-off between outlier detection (fairness) and outlier resistance (robustness). Positive tilting magnifies outlier losses and thus excels when rare points are critical to the task. Negative tilting downweights these losses, yielding a more conservative classifier resilient to extreme noise. The consistent qualitative behavior observed across all three TERM optimizers highlights that this effect stems from the tilted weighting mechanism itself rather than any particular optimization algorithm. In online scenarios where data arrive sequentially, this simple modification restores the flexibility of TERM, enabling practitioners to smoothly navigate the robustness-sensitivity spectrum with a single hyperparameter.

\section{Discussion}

In practice, the choice of the tilt sign should reflect the role that high-loss examples play within the learning objective, which explains why the optimal direction differs between the two toy problems considered. For online robust linear regression, the goal is to recover the true slope and intercept of the dominant inlier distribution while treating occasional extreme points as corruptions. In this context, large residuals are nuisance signals; points far from the current fit are likely outliers or noise, and thus their gradient influence should be diminished. Negative tilting achieves this by exponentially downweighting large losses, effectively clipping gradients and enabling the estimator to focus on the majority of inliers. Empirically, this directs the parameter vector steadily toward the true coefficients and maintains stability once convergence is reached.

Conversely, in the streaming binary classification task aimed at outlier detection, rare points constitute the positive class and are the primary targets for identification. Large losses correspond to misclassified outliers, necessitating aggressive updates to the decision boundary to better capture such points in future iterations. Positive tilting magnifies these losses, sharply increasing gradient contributions and pulling the separating hyperplane toward outliers. Negative tilt, by contrast, suppresses this crucial information, resulting in a conservative classifier that seldom flags outliers. Therefore, negative tilt is preferable when robustness to outliers is desired (as in regression), while positive tilt is better suited for heightened sensitivity and outlier detection (as in classification). This exemplifies how the same tilted weighting mechanism can serve distinct statistical objectives depending on the task at hand.

Looking forward, a critical direction for further research is developing principled methods to automatically select the appropriate tilt parameter based on task characteristics or data properties. Since the optimal sign and magnitude of tilting depend heavily on the role of high-loss examples within the learning goal—whether to emphasize or suppress them—automated tuning strategies would greatly enhance the practical usability of online TERM. Such methods might leverage adaptive schemes, meta-learning, or data-driven heuristics to dynamically adjust tilting during training, ensuring robust and sensitive performance without manual intervention.

In summary, the flexibility afforded by the tilting parameter $t$ in online TERM provides a powerful tool to navigate the trade-off between robustness and sensitivity, but realizing its full potential requires further advances in automated hyperparameter selection tailored to diverse learning scenarios.

\section{Conclusion}

We introduced an online variant of the Tilted Empirical Risk Minimization (TERM) framework that preserves the tunable trade-off between robustness and fairness in strict streaming settings where only one sample is processed at a time. Our approach leverages an exponential weighting scheme to maintain tilt sensitivity without additional memory or computational burden, overcoming limitations of minibatch TERM at $N=1$.

Through extensive experiments on synthetic streaming regression and binary classification tasks with outliers, we demonstrated that the sign and magnitude of the tilt hyperparameter $t$ enable a smooth interpolation between standard ERM behavior ($t \to 0$), robustness to outliers ($t < 0$), and fairness-sensitive learning ($t > 0$). Notably, negative tilting significantly improved convergence stability and accuracy in robust regression, while positive tilting enhanced recall for outlier detection in classification.

These findings highlight online TERM as a practical and theoretically principled approach for optimization that adapts seamlessly to diverse robustness and fairness requirements with minimal tuning. Future work will investigate adaptive schemes for automatically selecting tilt parameters tailored to specific tasks and data distributions, further expanding TERM’s applicability in real-world streaming applications.

\bibliographystyle{IEEEbib}
\bibliography{refs}

\end{document}